\documentclass[11pt,a4paper]{article}
\usepackage{authblk}
\usepackage[hyperref]{emnlp2018}
\usepackage{times}
\usepackage{latexsym}
\usepackage{graphicx}
\usepackage{multirow}
\usepackage{url}
\usepackage{booktabs}

\usepackage{url}

\aclfinalcopy 

\title{Adversarial Neural Networks for Cross-lingual Sequence Tagging}

\author[1]{Heike Adel}
\author[2]{Anton Bryl}
\author[2]{David Weiss}
\author[2]{Aliaksei Severyn}
\affil[1]{Institute for Natural Language Processing (IMS), University of Stuttgart, Germany}
\affil[2]{Google AI}
\affil[ ]{\texttt{heike.adel@ims.uni-stuttgart.de}}
\affil[ ]{\texttt{[antonbryl$|$severyn$|$djweiss]@google.com}}

\date{}

\begin{document}
\maketitle
\begin{abstract}
  We study cross-lingual sequence tagging with little or no labeled data
  in the target language.  Adversarial training has
  previously been shown to be effective for training cross-lingual sentence
  classifiers. However, it is not clear if language-agnostic
  representations enforced by an adversarial language discriminator
  will also enable effective transfer for token-level prediction
  tasks.  Therefore, we experiment with different types of adversarial training
  on two tasks: dependency parsing and sentence compression.  We show that adversarial training
  consistently leads to improved cross-lingual performance on each
  task compared to a conventionally trained baseline.
\end{abstract}

\section{Introduction}
Cross-lingual modeling is especially interesting when generalizing
from a ``source'' language with labeled data to a ``target'' language
without fully supervised annotations.  While POS taggers
and dependency parsers are available in many languages
\cite{nivre2016universal}, data for tasks like sentence compression
\cite{Filippova2015} or classification \cite{chen2017,joty2017} 
is much harder to come by.  Past success for 
cross-lingual transfer is mainly achieved by label projection
\cite{tackstrom2013,wisniewski2014,agic2016}, which requires manual
efforts, machine translation or parallel data.

In this work, we address the challenge of building models that
learn cross-lingual regularities. Our goal is to avoid overfitting to
the source language and to generalize to new languages through the use of
adversarial training.  Ideally, 
our
models 
build internally language-agnostic
representations which can then be applied to any new target language.
Adversarial training has gained a lot of attention recently for domain
adaptation by building domain-independent feature
representations~\cite{ganin2016,chen2018}.  As we show in this work,
cross-lingual transfer can be treated as a language-specific
variant of domain adaptation.
This poses additional challenges to adversarial training:
In contrast to most domain
adaptation settings, the change from source to target involves not
only the word distributions changing (as when adapting models from
news to web data) but the entire vocabulary changing as well. To
address this, we use bilingual word embeddings
and universal POS tags as a common intermediate representation.
The second difficulty is that, to the best of our
knowledge, adversarial loss has only been applied to cross-lingual NLP
classification tasks ~\cite{chen2017,joty2017,chen2018} in which a single
output label is predicted.  Filling the gap, we are the first to show
that adversarial loss functions are also effective for cross-lingual sequence
tagging in which multiple outputs are predicted for a given input
sequence.

We show for both a syntactic task (dependency parsing)
and a semantic task (extractive sentence compression)
that adversarial training improves cross-lingual
transfer when little or no data is available. For completeness, we
also provide a negative result: for training POS taggers, 
bilingual word embeddings and adversarial training
are not sufficient to produce useful cross-lingual models.

Our
contributions are:
(i) We adapt adversarial training for structured prediction and compare
  gradient reversal \cite{ganin2016}, GAN \cite{goodfellow2014} and
  WGAN \cite{arjovsky2017} objectives.
(ii) We show that this procedure is useful for both a syntactic 
and a semantic task.

\section{Related Work}
\label{sec:relatedWork}
Adversarial training \cite{goodfellow2014}, is receiving increased interest in the  NLP
community
\cite{Gulrajani2017,Hjelm2017,Li2017,Press2017,Rajeswar2017,Yu2017,Zhao2017}.

\newcite{ganin2016} propose adversarial training
with a gradient
reversal layer for
domain adaptation (for image classification and
sentiment analysis, resp.).  Similarly, \newcite{chen2017} and
\newcite{joty2017} apply adversarial training to cross-lingual
sentiment classification and community question answering,
respectively.  While most previous work in NLP has investigated
adversarial domain adaptation for sentence-level classification tasks,
we are the first to explore it for cross-lingual sequence tagging. Moreover, we
provide a direct comparison of different adversarial loss functions in
a cross-lingual training setting. 
\newcite{li2016} also work on adversarial sequence tagging but
treat sequence tag prediction and sequence labeling as adversarial model parts.
This is very different in motivation than using adversarial training 
in cross-lingual settings.
More related to our work is the
paper by \newcite{yasunaga2018} which applies adversarial training
in the context of part-of-speech tagging.  However, their model is
different from ours in two crucial respects.  They train
single-language models whereas we train a cross-lingual model.  They
use adversarial training in the form of input
perturbations whereas we use adversarial loss functions in order to
derive language-independent representations to be able to transfer
knowledge between source and target languages.

\section{Model}
We implement our model using DRAGNN, a Tensorflow framework
for efficient training 
of recurrent neural networks \cite{dragnn}.

\subsection{Model Architecture}
\label{sec:featureGen}
\begin{figure}
\centering
\includegraphics[width=.49\textwidth]{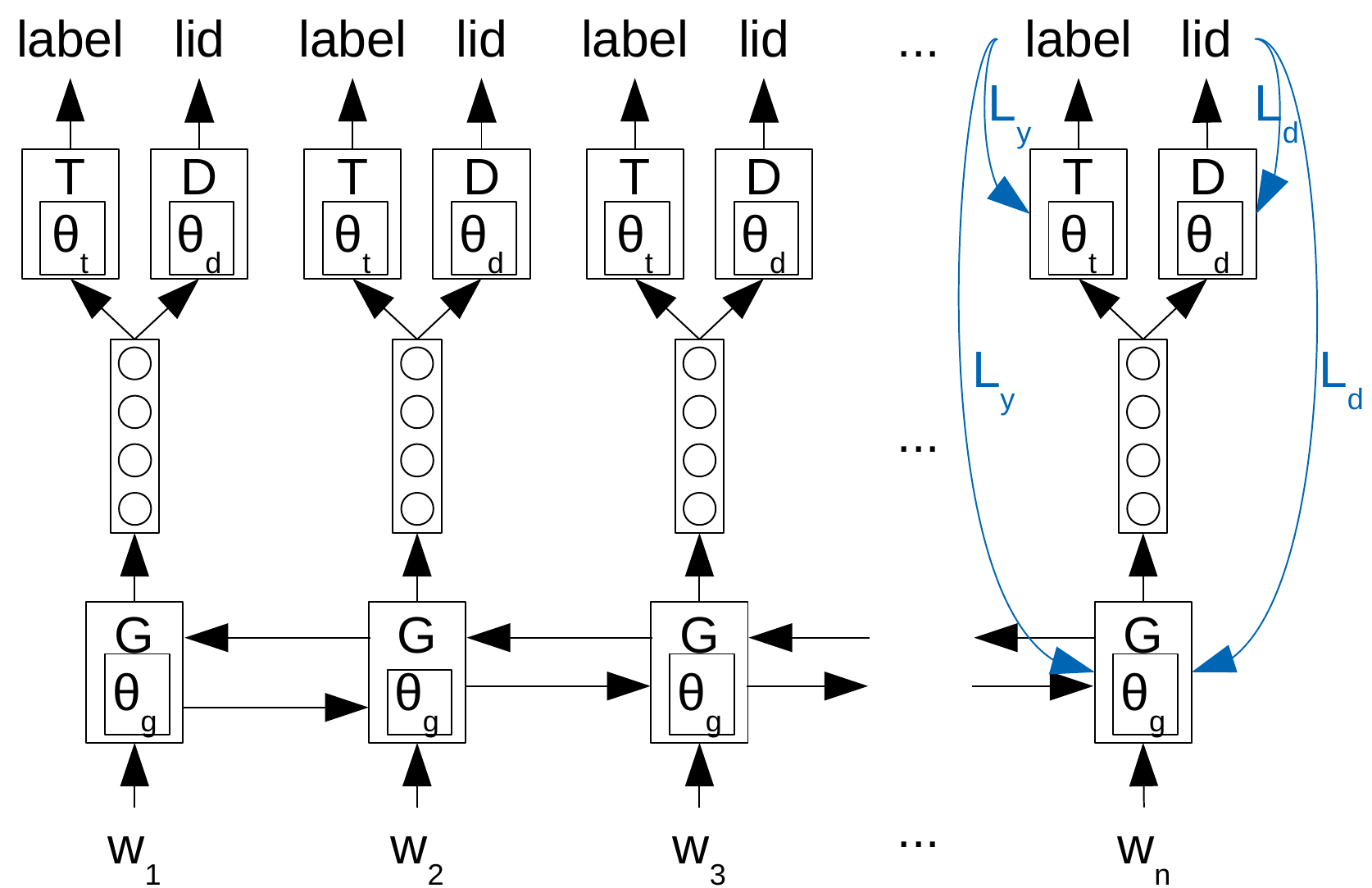}
\caption{General architecture of our model. G: feature generator 
(bi-LSTM),
T: target sequence tagger, D: domain discriminator. On the right, we illustrate
the flow of the different losses.}
\label{fig:generalModel}
\end{figure}

Our model consists of three main components:
a feature generator, a domain discriminator and
a sequence tagger: see 
Figure \ref{fig:generalModel}.

The \emph{feature generator} $G$ is a bi-directional LSTM (bi-LSTM) with one layer \cite{Hochreiter1997}
which uses the embeddings of words, their POS tags and Brown clusters as input (see Section \ref{sec:experiments} for more details).
Its output is consumed by the \emph{domain discriminator} $D$ and the \emph{target
tagger} $T$, which are both implemented as feed-forward networks and predict 
the language id (lid) of the input sequence and the target token label at each step, respectively.\footnote{For dependency parsing, the tagger is replaced by the arc-standard model from \cite{dragnn} which predicts two labels per token, but the feature generator architecture is the same.}
We found that predicting language id at the token level was more effective than predicting it on the sentence level.
The tagger objective maximizes the log-likelihood of the target tag sequence:
\begin{equation}
\label{eq:tagger}
\small
O_t = \max_{\theta_t} \sum_{i = 1}^{|y|} \log p(y_i | y_{<i}; G(x)) \label{crossentropy}
\end{equation}
updating w.r.t. parameters of the tagger $\theta_t$.
The objectives for the discriminator and feature generator depend on the adversarial techniques, which we describe next.

\subsection{Training with Gradient Reversal}
The first adversarial architecture we investigate
is gradient reversal training as proposed by \newcite{ganin2016}.
In this setting, the discriminator is a classifier, which
identifies the input domain given a single feature vector $x$.
Thus, its objective is $O_d = \max_{\theta_d} \log p(lid(x) | G(x))$.

The goal of the generator is to fool the discriminator which is achieved by updating the generator weights in the opposite direction w.r.t. the discriminator gradient:
\begin{equation}
\small
\theta_g := \theta_g + \nabla_{\theta_g} O_t - \lambda \nabla_{\theta_g} O_d
\end{equation} 
where $\lambda$ is used to scale the gradient from the discriminator.

\subsection{Training with GAN and WGAN}
The other two adversarial training objectives we investigate
is the GAN 
\cite{goodfellow2014} and its variant Wasserstein GAN (WGAN)
objective \cite{arjovsky2017}.
In contrast to gradient reversal, the discriminator inputs are sampled from source and target distributions $p_s$ and $p_t$, resp.
The adversarial objective for GAN is:
\begin{equation}
\label{eq:gan}
\small
O_d = \max_{\theta_d} E_{x \sim p_t}[\log D(G(x)] + E_{x \sim p_s}[\log (1 - D(G(x))]
\end{equation}
The objective of the feature generator is to act as an adversary w.r.t. the discriminator while being collaborative w.r.t. the target tagger: 
\begin{equation}
\small
O_g = \max_{\theta_g} [O_t - O_d]
\label{eq:generator}
\end{equation}

To stabilize the adversarial training, \newcite{arjovsky2017} have proposed WGAN, which trains the discriminator as:
\begin{equation}
\label{eq:wgan}
\small
O_d = \max_{\theta_d \in L} E_{x \sim p_t}[D(G(x)] - E_{x \sim p_s}[D(G(x_s)]
\end{equation}
with restricting the range of the discriminator weights for Lipschitz continuity.
The objective for the feature generator has the same form as in Eq.~\ref{eq:generator}.

Thus, the feature generator is incentivized to extract language-agnostic representations in order to fool the discriminator while helping the tagger.
As a result, the tagger is forced to rely more on language-agnostic features. 
As we discuss later, this works well 
for higher-level syntactic or semantic tasks
but does not work for low-level, highly lexicalized
tasks like POS tagging.

Note that the updates to the target tagger are affected by the discriminator only indirectly through the shared feature generator. 

\section{Experiments}
\label{sec:experiments}
We address two cross-lingual sequential prediction tasks:
dependency parsing and extractive sentence compression.
For both tasks, we evaluate four different settings:
(i) No ADA: no adversarial training:
the feature-generator and tagger are trained on 
the source language and then tested on the target language, 
(ii) GR: gradient-reversal training, (iii) GAN: using GAN loss,
(iv): WGAN: using WGAN loss. In (ii)-(iv), a discriminator is trained in
order to achieve language-agnostic representations in the
feature generator. 
In all setups, we use bilingual word embeddings (BWE),
Brown clusters
and universal POS tags as
input representation that is common across
languages. 
The BWEs are trained on unsupervised multi-lingual corpora
as described in \newcite{soricut2016}.
The POS tags are predicted by simple bi-LSTM taggers trained on the full datasets.

\subsection{Data and Evaluation Measure}
For parsing, we use
the French (FR) and Spanish (ES) parts of the Universal Dependencies v1.3
\cite{nivre2016universal}. 
For sentence compression, in the absence of non-english datasets,
we collect our own datasets for FR and ES 
from online sources (e.g., 
Wikipedia) and ask professionally trained linguists to label 
each token with KEPT or DROPPED (see Section \ref{sec:sc})
such that the compressed sentences 
are grammatical and informative.
See Table \ref{tab:stat} for statistics.
In all our experiments, we use ES as source and FR as target.

We follow standard approaches for evaluation: 
token-level labeled
attachment score (LAS) for dependency parsing and
sentence-level accuracy for sentence compression \cite{Filippova2015}.
Note that for the latter, 
all token-level KEPT/DROPPED decisions need to be correct
in order to get a positive score for a sentence.

Note that, for all experiments without target training data,
we do not use target data for tuning the models. Thus,
the sequence-tagger is fully unsupervised w.r.t. the target language.

\begin{table}
\centering
\scriptsize
\begin{tabular}{lcc}
 & Spanish & French\\ 
 \midrule
\#sent. (train / dev / test) & 2353 / 115 / 115 & 2760 / 115 / 115\\
avg sent. length & 41 tokens &  37 tokens\\
compression rate & 26.8\% & 37.8\%\\ 
\end{tabular}
\caption{Statistics of SC dataset.}
\label{tab:stat}
\end{table}

\begin{table}[]
\centering
\footnotesize
\begin{tabular}{lllll}
  Training data          & No ADA & GR    & GAN            & WGAN           \\ 
\midrule
ES + 0k FR & 63.35 & {\bf 64.25} & 61.20 & 62.51 \\ 
ES + 1k FR & 67.33 & {\bf 67.43} & 66.69 & 66.97      \\ 
ES + 2k FR & 68.51 &{\bf 69.05} & 68.54 & 68.14 \\ 
ES + all FR & 80.17 & 80.46 & {\bf 80.68} & 80.22 \\
\end{tabular}
\caption{es$\rightarrow$fr dependency parsing results.}
\label{parsing}
\end{table}

\subsection{Dependency Parsing}
We train the tagger on
ES data with and without adversarial loss.  Table \ref{parsing}
shows the results. Adversarial training gives consistent improvements
over the conventionally trained baseline in all settings. 
It also outperforms a monolingual model trained on the full dataset of the 
target language FR which achieves a score of 80.21.
When comparing the different adversarial loss functions,
GR outperforms GAN and WGAN in most cases. 
One possible reason is a difference in the discriminator's strength:
During training,
we observe that the discriminator of GAN and WGAN could easily predict the language id correctly (even after 
careful tuning of the update rate between generator and discriminator).
This is a well-known problem with training GANs: When the discriminator becomes 
too strong it provides no useful signal for the feature generator. 
In contrast, GR training updates the feature generator by taking
the inverse of the gradient of the discriminator cross-entropy loss. 
This simpler setup possibly results in a better training signal for the generator.

\begin{table}
\centering
\footnotesize
\begin{tabular}{lrr}
Training data                    & Accuracy      \\ 
\midrule
1k FR          & 0.85            \\ 
1k MT-FR          & 1.71          \\ 
1k FR + 1k MT-FR  & 17.09        \\ 
2k FR          & 23.93      \\ 
All FR         & 25.64   \\ 
\end{tabular}
\caption{Monolingual SC results on French.}
\label{tab:SCmono}
\end{table}

\begin{table}[]
\centering
\footnotesize
\begin{tabular}{lrrrr}
 Training data          & No ADA & GR    & GAN            & WGAN \\ 
\midrule
ES + 0k FR & 0.00  & \textbf{9.17} &   0.00     & 1.71 \\ 
ES + 1k FR &  20.51 & \textbf{22.27} & 11.97 &   15.38 \\ 
ES + 2k FR & \textbf{29.06} & 24.89 & 15.38 & 19.66 \\
ES + all FR & 29.91 & \textbf{30.77}  & 22.22 & 29.06 \\
\end{tabular}
\caption{es$\rightarrow$fr SC with adversarial training.}
\label{tab:SCada}
\end{table}

\subsection{Sentence Compression}
\label{sec:sc}
Extractive sentence compression aims at generating shorter versions of a given sentence by
deleting tokens \cite{Knight2000,Clarke2008,Berg-Kirkpatrick2011,Filippova2015,Klerke2016}.
This is useful for text summarization as well as for
simplifying sentences or providing shorter answers to questions.
We follow related work and treat it as a sequence-tagging problem: 
each token of the input sentence is tagged with either
KEPT or DROPPED, 
indicating which words should occur in the compressed sentence. 
To solve the task, the model needs to consider the meaning of
words and sentences. Thus, we consider it a semantic task although
we treat it as a sequence-tagging problem.

\textbf{Monolingual and MT models.}
Most work on sentence compression considers English corpora only.
Studies on other languages either train different monolingual models \cite{Steinberger2007}
or use translation or alignments to transfer compressions from English into another language \cite{Aziz2012,Takeno2015,Ive2016}.
In order to get baseline results, we follow these approaches 
and train monolingual models (on FR) and models on translated 
data (MT-FR).\footnote{We use the Google MT API to translate from ES into FR.}
Thus, the feature generator and tagger are
monolingual models
and there is no language discriminator.
Table \ref{tab:SCmono} shows the results. 
We find that MT can help to train first models in a new language. However,
training data in the target language is better (see performance
gap from 1k FR+1k MT-FR to 2k FR).

\textbf{Cross-lingual models.}
Next, we train cross-lingual models (see Table \ref{tab:SCada}).
Even without adversarial training, the models perform better than the monolingual models. This shows that
information can already be shared from ES to FR by using bilingual word embeddings.
When adding adversarial training, we again notice a better performance of GR
compared to GAN or WGAN: GR training boosts the results, especially for no or little
FR training data. GAN and WGAN loss does not perform as good as for dependency parsing.

\section{Discussion}
We compared tasks of different natures:
a syntactic task (depedency parsing)
and a semantic task (sentence compression).
For completeness, we also report that language-agnostic
POS taggers did not lead
to promising results:
Even though adversarial training improved a No-ADA baseline by
several points, 
cross-lingual transfer still yielded only 45\%
POS accuracy in the target language, which is not accurate enough to be
useful 
in downstream models.
We assume that POS tagging requires
seeing language-specific vocabulary more than other tasks.
However, 
we showed that the more high-level tasks dependency parsing 
and sentence compression can benefit from language-agnostic
representations. 
To the best of our knowledge, this is the first work
showing the effectiveness of adversarial loss
for cross-lingual sequence tagging.
Therefore, we opted for a language pair from the same
language family. 
However,
our algorithm is applicable to language pairs
from different language groups as well. 

\section{Conclusion and Future Work}
In this paper, we study the utility of adversarial training for cross-lingual sequence-tagging.
Our results show that the more higher-level structure the task required, 
the more gains we could achieve with cross-lingual models.
Gradient reversal training outperformed GAN and WGAN loss in our experiments.
In future work we plan to extend our study to other language pairs, 
including languages from different families.


\bibliography{refs}
\bibliographystyle{acl_natbib_nourl}



\end{document}